# Gapoera: Application Programming Interface for AI Environment of Indonesian Board Game

Rian Adam Rajagede[1, a)] and Galang Prihadi Mahardhika[1, b)]

Author Affiliations
[1] *Department of Informatics, Universitas Islam Indonesia, Jl. Kaliurang km. 14.5, Sleman, 55584, Indonesia*

Author Emails
[a)] Corresponding author: rian.adam@uii.ac.id
[b)] galang.prihadi@uii.ac.id

**Abstract.** Currently, the development of computer games has shown a tremendous surge. The ease and speed of internet access today have also influenced the development of computer games, especially computer games that are played online. Internet technology has allowed computer games to be played in multiplayer mode. Interaction between players in a computer game can be built in several ways, one of which is by providing balanced opponents. Opponents can be developed using intelligent agents. On the other hand, research on developing intelligent agents is also growing rapidly. In computer game development, one of the easiest ways to measure the performance of an intelligent agent is to develop a virtual environment that allows the intelligent agent to interact with other players. In this research, we try to develop an intelligent agent and virtual environment for the board game. To be easily accessible, the intelligent agent and virtual environment are then developed into an Application Programming Interface (API) service called Gapoera API. The Gapoera API service that is built is expected to help game developers develop a game without having to think much about the artificial intelligence that will be embedded in the game. This service provides a basic multilevel intelligent agent that can provide users with playing board games commonly played in Indonesia. Although the Gapoera API can be used for various types of games, in this paper, we will focus on the discussion on a popular traditional board game in Indonesia, namely Mancala. The test results conclude that the multilevel agent concept developed has worked as expected. On the other hand, the development of the Gapoera API service has also been successfully accessed on several game platforms.

## INTRODUCTION

The development of computer games is accelerating nowadays, particularly those played online. Computer games can now be played in a multiplayer mode thanks to advancements in Internet technology. In a computer game, the interaction between players can be achieved in a variety of ways, one of which is by providing intelligent agents as opponents in the game. Research in intelligent agents today is growing rapidly, especially after the popularity of Deep Learning methods in 2012 [1]. It has affected many areas of software development, including the development of games. In game development, research related to the development of intelligent agents that utilizes the Deep Learning method is carried out to facilitate single-player and multiplayer games [2]–[4]. The development of intelligent agents in a game is quite popular because the game interaction is considered to simulate agent interaction in other domains but in a cheaper way.

In various types of games, human-level abilities have been surpassed by intelligent agents, such as in Mancala [5], [6], Chess [7], [8], or Go [3]. One of the popular algorithms used in agent development is deep reinforcement learning that combines deep neural networks and reinforcement learning. Deep reinforcement learning works by training agents to choose an action based on the best value obtained each time the agent plays (the more agents play, the better the action chosen by the agent) [2]. In this way, agents will learn from their interactions with the game to determine which best move should be chosen in a given condition.

When developing agents in a game, researchers need a virtual environment for agents to interact with the game. The interaction of agents with games is crucial in training and evaluation to understand the consequences of each action. This interaction can be in the form of interactions with objects in the game or interactions with other players who are opponents or teams. Therefore, it is necessary to develop a good environment to support agent development. One environment that is quite often used is Open AI Gym [9]. Gym provides many single-agent environments ranging from single-player games to classic controls (including the Atari 2600 [2]). As for board games, two environments for reinforcement learning that are quite popular are OpenSpiel [10] and PettingZoo [11].

Indonesia is a country that has a diverse culture, including traditional board games. One of the games in Indonesia that is quite popular is Mancala (in Indonesia known as Congklak), which is thought to have come from Africa or Saudi Arabia [12]. In Indonesia, Mancala has a variety of names and variations of rules that differ in each region. Unfortunately, the development of digital board games in Indonesia, both for Mancala and other games, is still not centralized and organized. Researchers are often required to repeat creating the game environment whenever they start to develop games. In the previous research on game agents for Mancala, the development code can not be freely accessed by the public [5], [6], [13], both the agent code and the environment. Another study made a digital version of Mancala [14], which still does not provide public access to its development code. On Github, one of the Mancala projects that are still active is carried out by Talebi [15]. The project builds the Mancala API using Spring Boot, but even though the code is publicly available, users have to build the API in their own environment. By publicly providing the environment code and access, the researcher not only can focus more on the intelligent agents but the environment can also be used as a standard or benchmark among researchers.

In this paper, we introduce an API that we named Gapoera API. Gapoera API is an API-formed gaming environment that includes intelligent agents. Gapoera API is developed using the general board game designs and has run two traditional Indonesian games, Mancala and Surakarta Chess. This API can later be utilized from two points of view. This API can be used for AI researchers to develop intelligent agents, especially for Indonesian board games. The provided intelligent agents can be a benchmark for measuring the agent's performance made by another researcher. Additionally, this API can also be used as a game engine that game developers can use to develop board games in general. With features such as multilevel agents, human players do not have to fight the agent who could surpass humans' ability. Players can choose agents with levels that match their abilities.

## METHOD

### Mancala Game

Mancala Game is a popular board game that is played almost everywhere in the world [16]. This game has many variations and different names in each region. In Indonesia, this game is quite popular under the name Congklak or Dakon, while several other names for this game are Congkak in Malaysia, Bao in Africa, and Kalah in America. The rules of the game and the details of the components in each area can be different.

This study implements one of the well-documented Mancala rules known as Kalah variant rules [17]. One of the rules that need to be highlighted from this variant is that if the player puts his final stone in his home-pit (the rightest pit), the players get an extra turn. We use parameterized mancala board size so that we can set the number of pits and stones in each pit. By default, in this study, we use Mancala with a variance of seven pits and seven stones in each pit or called Mancala (7,7). The number of pits and stones in each pit was chosen because this form is popular in Indonesia. Above that, the smaller size of Mancala can also be considered as solved games [18] (a game that the winner can be determined from the start if both players play optimally)

### Board Game Analysis

Board games have different characteristics compared to single-player video games previously researched [2]. In most board games, the agent must interact with one or more players. This environment behavior is called a Markov Game [19]. The process of interaction between the agent, the environment, and the opponent is shown in Fig. FIGURE **1**.

In general, at a timestep t, the agent will receive a state *S* representing the board's condition and then perform one of the legal actions *A*. The selection of this action is based on the algorithm used by agents. In the next turn, the opponent will take action. The action from opponents can be seen as an effect that changes the state outside of our control. Their action will make the game reach a new state *S'* in our agent's next turn. The agent will get two types of rewards, *R*, the immediate reward obtained after performing an action, and the long-term reward obtained if the agent wins the match. We can see a game as a state sequence or called the Markov Decision Process (MDP from these components). For example, in some algorithms, reinforcement learning, the agent learns from a collection of experiences represented by the tuple of (*S*, *A*, *R*, *S'*).

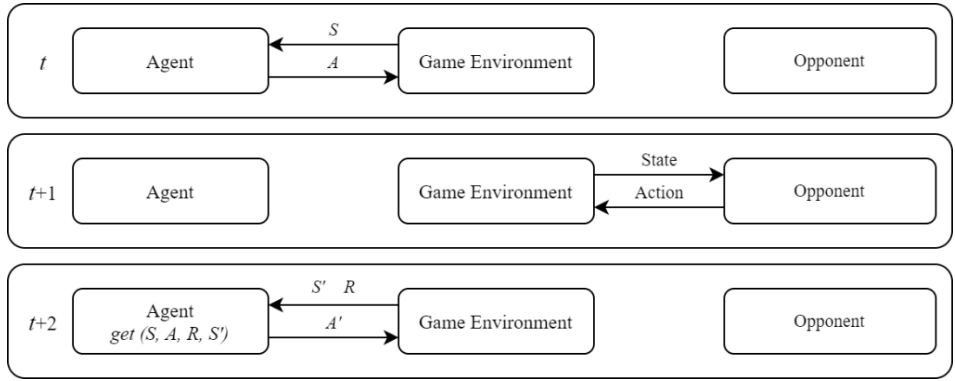

**FIGURE 1.** Agent-environment interaction in turn-based game.

In some studies, there is also a term the observation. Just like the state, the observation stores information that occurs at a time in the game, but the observation shows information that is only visible from the viewpoint of an agent. In perfect information games such as Mancala and Chess, the observation state and state contain the same information. However, in some types of games, such as card games, the game state displays more general information, and observation will display specific information from the point of view of one player, such as cards owned (not cards owned by other players).

By using a framework (*S*, *A*, *R*, *S'*), we can implement an API environment that can run for a wide variety of games. Each game environment must be able to define what state (*S*), what actions can be performed (*A*), rewards, and what new state is obtained (*R* and *S'*). From the above analysis, we can see that the Mancala game can be described as an MDP with the details of each component, as shown in TABLE 1.

**TABLE 1.** Mancala's state, action, and reward analysis

| Component | Description |
| --- | --- |
| State | Number of stones in all pits, both our pits and opponent's pits |
| Action | The number of pits that can be taken. For Mancala(7,7) the number of available action that can be taken for each player, is a maximum of seven. |
| Reward | There are two types of rewards used. For immediate rewards, we use the number of stones obtained in the player's home-pits. If the player wins the game, a winner reward will also be obtained. |

What distinguishes between the Mancala variant that we implemented compared to other games, for example chess, is that there is a rule where players can do more than one consecutive action. So that the change from one state *S* to the next state *S'* does not always occur after both players perform the same action. This is illustrated in Fig. FIGURE **2**.

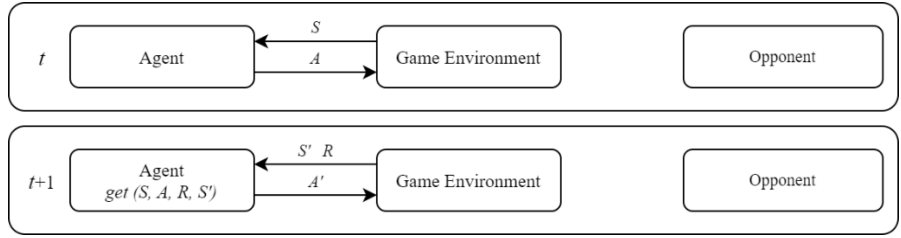

**FIGURE 2.** Agent-environment interaction in special case in Mancala game

## Intelligent Agents

In this study, we focused on the use of AI agent as a player in the game. Agents will act as opponents who interact with other players and the environment. Determining a good opponent to play is very important to achieve the goal of playing itself. The level of difficulty of the opponent plays a major role in determining player engagement in the game [20]. Whereas in artificial intelligence research, in which the game generally tries to find the most powerful agent, the opponent must be adequately competitive and represents a strong opponent.

In turn-based games, the algorithm that is commonly used to create intelligent agents is using the searching method. The agent searches for possible states that can occur up to n- steps ahead when a sequence of actions is taken. The agent will choose the best action that leads to victory. Some commonly used search algorithm is a Monte-Carlo Tree Search [21] and α-β Pruning [22]. The search method that combines a deep neural network to decide the action has been carried out on Go and Chess [3], [7]. Another approach is to predict which action should be taken in a state without searching. This method is actually better suited to be done by a single-player game [2]. In [23], researchers have also compared the agent performance that uses the search method with agents that do not use the search method in the Othello game. The result shows that an agent that predicts an action without searching and only using the Convolutional Neural Network (CNN) is still able to compete against agents that search up to 2 steps ahead.

In this study, we put multilevel agents as a feature in the Mancala game in the Gapoera API. Multilevel agent means agents have different levels of difficulty. We created two agents based on two simple strategies commonly used in Mancala. We gave the agent's names as Greedy Agent I and Greedy Agent II, as explained in TABLE 2.

**TABLE 2.** Mancala agents' specification

|  | **Greedy Agent I** | **Greedy Agent II** |
| --- | --- | --- |
| **Description** | Adding as many stones as possible to the home-pit | Getting as many extra turns as possible |
| **Steps** | 1. Take the pit that can add the most stones to the home pit. This includes when the stone that we put into our own empty pit will take the stone from the opponent's pit.<br>2. If there is more than one pit that meet the condition, choose randomly | 1. Take a pit that can add more turns.<br>2. If there is more than one pit, select the rightmost one. (if the rightmost pit is empty, then in the next turn, it can be filled with one stone and add more turn in the next round). |

The greedy agent will look at all valid actions and then choose an action based on the priority of his choice. In this process, the agent will simulate every action in the environment to see the reward and state obtained.

To ensure that the agent does not perform the same action all the time, the agent has some random element in choosing the action. First, when there are several actions that have the same profit value, the agent will choose randomly. Second, the agent will be given an exploration parameter ε with a value of *p,* which is not equal to 0 (ε = *p*; *p* ≠ 0 ), which means the agent will perform a random action with a probability of *p* regardless of the previous rules.

# RESULTS

## Gapoera API: Indonesian Board Game Environment

In this study, we developed the Gapoera API[1], an Indonesian Board Game Environment API, using Python and Flask[2]. Gapoera API provides a gaming environment and AI agent in the game that can be used both by AI researchers as well as game developers. Currently, there are two game environments that can be used, namely Mancala and Surakarta Chess. An illustration of using the Gapoera API is shown in Fig. 3.

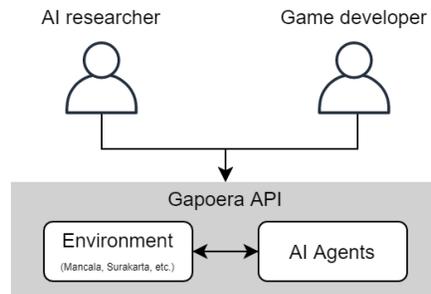

**FIGURE 3.** Illustration of Gapoera API usage

In general, users can do the following: 1. initializing the game, 2. take action, 3. Get some game information. The flow of using the API from game initialization to completion is shown in Fig. FIGURE 4. The first step is to initialize the game. The user will then get a unique game id. All information of an active game will be stored on the server indefinitely to prevent users from cheating on the game by taking illegal actions. The user can assign actions to the environment. If this action is valid, then the game state will change. In Fig. FIGURE 4, action selection is carried out in the box with the dotted line; this is to indicate that this process occurs on the client-side

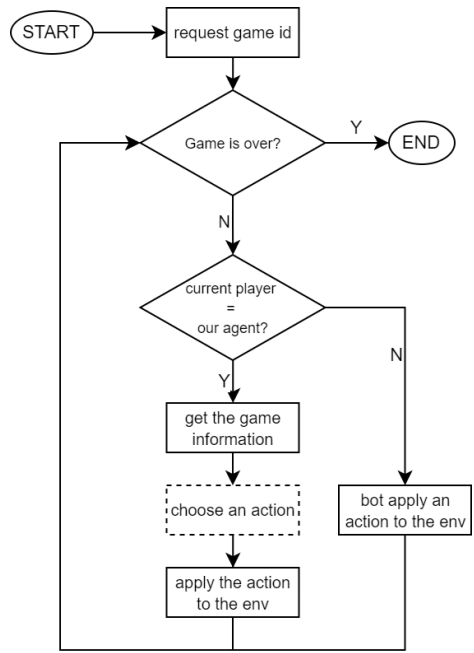

**FIGURE 4.** Flowchart of Gapoera API usage in a turn-based game. The box with the dotted line is a process on the client-side

---

[1] http://api.gapoera.com
[2] https://flask.palletsprojects.com/en/2.0.x/

*Basic usage in Python*

In this section, we try to show the use of the API in Python for readers who want to see its technical use. Generally, the use of the API for turn-based two-player games, such as Mancala, can be seen in Fig. **FIGURE 5**. The user can obtain the current game state via `game.state` (Fig. 5.a) or observation state using `game.observe` (Fig. 5.b).

In the next step, a user agent must determine what action needs to be taken. The choice of this action depends on the type of game being played. The `game.step` method is used to apply the selected action to the environment. In a game against a provided bot, the user asks the bot to act with `game.bot_step`.

```
1  game = Game(game_name, host_url)
2  game_id = game.start()
3
4  while game.is_over==False:
5      if game.current_player==0:
6          state = game.state
7          action = agent(state)
8          game = game.step(game_id, action)
9      else:
10         game = game.bot_step(game_id)
```
(a)

```
1  game = Game(game_name, host_url)
2  game_id = game.start()
3
4  while game.is_over==False:
5      if game.current_player==0:
6          observation = game.observe(game.current_player)
7          action = agent(observation)
8          game = game.step(game_id, action)
9      else:
10         game = game.bot_step(game_id)
```
(b)

**FIGURE 5.** Basic API usage that is written in Python

*Simulation*

The Greedy Agent must be able to consider every action based on the reward gained from each action. The Gapoera API allows users to perform multi-step simulations in the environment used. In the simulation, the sequence of actions performed does not change the actual environment. We implement a memory stack for saving the game state so that the agent can create a game tree during simulation. The simulation starts when the user runs the `game.sim_start` and stops when the user runs `game.sim_stop`.

*Multiplatform Game Development*

With the form of an API, users can use it for various needs. Researchers who want to research intelligent agents can call this API regardless of the programming language as long as they are connected to the internet. Although developing an interface is not required to use the API, we created two Mancala games on different platforms when testing the system. We made Mancala game using Python Tkinter[3] (Fig. **FIGURE 6**) and Construct 3[4] (Fig. **FIGURE 7**).

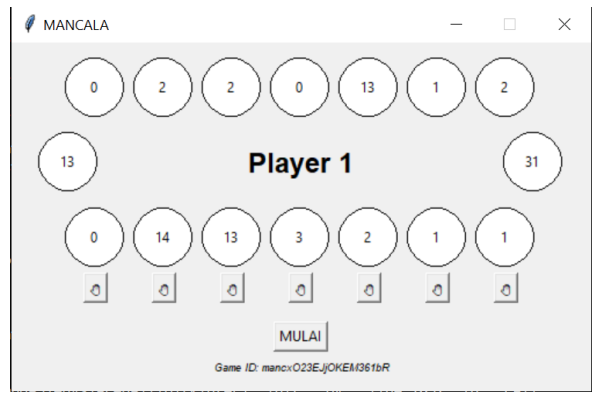

**FIGURE 6.** Mancala API used in desktop implemented using Python Tkinter

---

[3] https://docs.python.org/3/library/tkinter.html
[4] https://www.construct.net

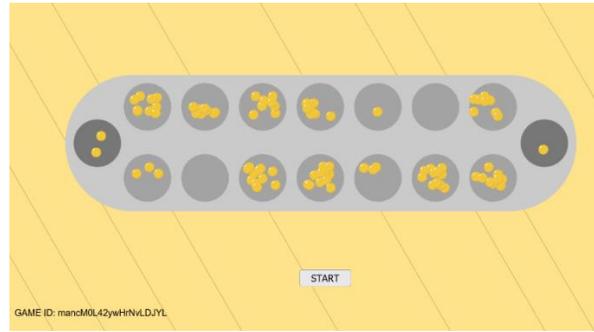

**FIGURE 7.** Mancala API implemented using Construct 3 game engine

## Agents Evaluation

We tested the Greedy Agents (GA I and GA II) described previously to determine the agent levels. We made two different versions for each agent with a value of ε = 0.1 and ε = 0.3. Both agents played in ten matches. Each agent will take turns being the first player to take action after five games. The results of the game are shown in Table TABLE 3. The table shows that GA I outperformed in every case even when ε = 0.3 or 30% of its action is random.

We define three levels of the agent provided in the API based on the total winnings obtained. From the most difficult level to the easiest is GA I (ε = 0.1) with a total of 15 wins from 20 matches, GA I (ε = 0.3) with a total of 12 victories, and GA II (ε = 0.1) with a total of seven victories.

**TABLE 3.** The result of ten games between Greedy Agent I (GA I) against Greedy Agent II (GA II)

|  | GA II (ε = 0.1) | GA II (ε = 0.3) |
|---|---|---|
| **GA I (ε = 0.1)** | GA I win 7 games<br>GA II win 3 games | GA I win 8 games<br>GA II win 2 games |
| **GA I (ε = 0.3)** | GA I win 5 games<br>GA II win 4 games<br>Tie 1 game | GA I win 7 games<br>GA II win 2 games<br>Tie 1 game |

## CONCLUSION

In this study, we developed the Gapoera API, a game environment that can be used in general, either by AI researchers or game developers. AI researchers can use the Gapoera API with intelligent agents to develop intelligent agents in the Indonesian board game environment. The multilevel intelligent agents available can serve as benchmarks or rivals when training models. With various levels available, human players can also enjoy the game in Gapoera API by fighting agents with different levels. For future research, we will try to improve the quality of currently available agents by involving human capabilities as a reference for development. Through this step, we hope to create agents who are not only strong but also can maintain better player engagement. From a system development point of view, Gapoera API development will also be improved in terms of security and reliability, especially when dealing with traffic when using this service.